\documentclass[11pt,a4paper]{article}
\usepackage[hyperref]{acl2017}
\usepackage{times}
\usepackage{latexsym}

\usepackage{url}

\usepackage{enumitem}
\usepackage{todonotes}
\usepackage{microtype}
\usepackage{booktabs}
\usepackage{amsmath, amssymb}
\usepackage{xparse}
\usepackage{cleveref}
\usepackage{multirow}
\usepackage[nolist]{acronym}

\usepackage[tight,footnotesize]{subfigure}

\newcommand{\phrase}[1]{{\sf #1}}
\newcommand{\triple}[1]{\texttt{(#1)}}
\newcommand{\resource}[1]{\texttt{#1}}
\newcommand{\literal}[2]{\texttt{"#1"\textasciicircum\textasciicircum<#2>}}
\newcommand{\literalPrefixed}[2]{\texttt{"#1"\textasciicircum\textasciicircum#2}}
\newcommand{\rdfLangString}{\texttt{rdf:langString}}
\newcommand{\class}[1]{\texttt{#1}}
\newcommand{\property}[1]{\texttt{#1}}
\newcommand{\individual}[1]{\texttt{#1}}

\newcommand{\realize}[1]{$\ensuremath{\rho(\texttt{#1})}$}

\newcommand{\pos}[1]{\texttt{pos}(#1)}

\usepackage{fixme}
\fxsetup{
    author=,
    layout=inline, 
    theme=color
}
\FXRegisterAuthor{lb}{alb}{LB}

\usepackage{xcolor}       
\definecolor{dkgreen}{rgb}{0,0.6,0}
\definecolor{gray}{rgb}{0.5,0.5,0.5}
\definecolor{light-gray}{rgb}{0.97,0.97,0.97}

\usepackage{tikz}
\usepackage{tikz-qtree}

\usepackage{listings}
\definecolor{olivegreen}{rgb}{0.2,0.8,0.5}
\definecolor{grey}{rgb}{0.5,0.5,0.5}
\lstdefinelanguage{ttl}{
  basicstyle=\small\ttfamily,
  sensitive=true,
  morecomment=[l][\color{grey}]{@},
  morecomment=[l][\color{olivegreen}]{\#},
  morestring=[b][\color{blue}]\",
}
\lstdefinelanguage{manchester}
{morekeywords={owl,xml,dc,rdf,skos,description,PlainLiteral,int,float,
        some,only,value,min,exactly,max,and,or,not,SOME,ONLY,VALUE,MIN,EXACTLY,MAX,AND,OR,NOT,
        Prefix,Ontology,Import,Individual,Facts,Types,Class,
        DataProperty,ObjectProperty,AnnotationProperty,Annotations,
DifferentIndividuals,SubClassOf,EquivalentTo,DisjointWith,DisjointUnionOf,SubPropertyOf,DisjointClasses,DisjointProperties,
Symmetric,Asymmetric,Reflexive,Irreflexive,Transitive,Functional,InverseFunctional,
        Characteristics,Range,Domain,Datatype},
     basicstyle=\small\ttfamily,
     keywordstyle=\bfseries,
     commentstyle=\color{gray},
     stringstyle=\color{blue},
     numbers=left,
     numberstyle=\tiny\color{gray},
     stepnumber=1,
     numbersep=10pt,
     tabsize=2,
     showspaces=false,
     showstringspaces=false,
     breaklines=true,                           
     sensitive=true,                            
     morecomment=[l][commentstyle]{\#},         
     morestring=[b]",                           
     numbers=none
}
\lstdefinelanguage{sparql}{%
   morekeywords=[1]{CONSTRUCT,WHERE,SELECT},
   morekeywords=[2]{AND,FILTER,UNION,OPT,OPTIONAL,MINUS,ORDER,GROUP,BY,DESC,OFFSET,LIMIT},%
   morekeywords=[3]{sameTerm,isBLANK,isLITERAL,isIRI,BOUND,DISTINCT},
   morekeywords=[4]{rdf,rdfs,owl,dbo,res,xsd},
   morekeywords=[5]{>},
   morestring=[b]",%
   alsodigit={-},%
}[keywords,strings]

\usepackage{colortbl}
\usepackage{amsmath}
\usepackage{xcolor}
\usepackage{graphicx}
\usepackage{array}
\usepackage{makecell}

\colorlet{tableheadcolor}{gray!25} 
\colorlet{tablerowcolor}{gray!10} 
\newcommand{\rowcol}{\rowcolor{tablerowcolor}} %

\lstdefinestyle{rdf}{numberblanklines=false, morekeywords={},
backgroundcolor=\color{backcolour},   
    commentstyle=\color{codegreen},
    keywordstyle=\color{magenta},
    numberstyle=\tiny\color{codegray},
    stringstyle=\color{codeblue},
    basicstyle=\footnotesize,
    breakatwhitespace=false,         
    breaklines=true,                 
    captionpos=b,                    
    keepspaces=true,                 
    numbers=left,                    
    numbersep=5pt,                  
    showspaces=false,                
    showstringspaces=false,
    showtabs=false,                  
    tabsize=2
}

\newcolumntype{R}[2]{%
    >{\adjustbox{angle=#1,lap=\width-(#2)}\bgroup}%
    l%
    <{\egroup}%
}

\lstdefinestyle{sparql}{numberblanklines=false, morekeywords={SERVICE,SELECT,DISTINCT,SAMPLE,FROM,WHERE,FILTER,ORDER,GROUP,BY,IN,AS,LIMIT}}

\title{A Holistic Natural Language Generation Framework for the Semantic Web}

\author{Axel-Cyrille Ngonga Ngomo$^{1}$ \, Diego Moussallem$^{1}$ \, Lorenz B\"uhmann$^{2}$ \\
         $^{1}$Data Science Group, University of Paderborn, Germany \\
         $^{2}$AKSW Research Group, University of Leipzig, Germany \\
         {\tt \{first.lastname\}@upb.de} \\
         {\tt lastname@informatik.uni-leipzig.de}}

\aclfinalcopy 

\date{}
\begin{document}

\begin{acronym}[UML]
	\acro{AOS}{Agricultural Ontology Services}
	\acro{AGRIS}{Agricultural Science and Technology}
	\acro{API}{Application Programming Interface}
	\acro{A2KB}{Annotation to Knowledge Base}
	\acro{BPSO}{Binary Particle-Swarm Optimization}
	\acro{BPMLOD}{Best Practices for Multilingual Linked Open Data}
	\acro{BFS}{Breadth-First-Search}
	\acro{CBD}{Concise Bounded Description}
	\acro{COG}{Content Oriented Guidelines}
	\acro{CSV}{Comma-Separated Values}
	\acro{CCR}{cross-document co-reference}
	\acro{DPSO}{Deterministic Particle-Swarm Optimization}
	\acro{DALY}{Disability Adjusted Life Year}
	\acro{D2KB}{Disambiguation to Knowledge Base}

	\acro{ER}{Entity Resolution}
	\acro{EM}{Expectation Maximization}
	\acro{EL}{Entity Linking}
	\acro{FAO}{Food and Agriculture Organization of the United Nations}
	\acro{GIS}{Geographic Information Systems}
	\acro{GHO}{Global Health Observatory}
	\acro{HDI}{Human Development Index}
	\acro{ICT}{Information and communication technologies}
    \acro{KB}{Knowledge Base}
	\acro{LR}  {Language Resource}
	\acro{LD}  {Linked Data}
	\acro{LLOD}  {Linguistic Linked Open Data}
	\acro{LIMES}{LInk discovery framework for MEtric Spaces}
	\acro{LS}  {Link Specifications}
	\acro{LDIF}{Linked Data Integration Framework}
	\acro{LGD} {LinkedGeoData}
	\acro{LOD} {Linked Open Data}
	\acro{MSE}{Mean Squared Error}
	\acro{MWE}{Multiword Expressions}
	\acro{NIF}{NLP Interchange Format}
	\acro{NIF4OGGD}{NLP Interchange Format for Open German Governmental Data}
	\acro{NLP}{Natural Language Processing}
	\acro{NER}{Named Entity Recognition}
	\acro{NED}{Named Entity Disambiguation}
	\acro{NEL}{Named Entity Linking}
	\acro{NN}{Neural Network}
	\acro{NLG}{Natural Language Generation}
	\acro{NN}{Neural Network}
	\acro{NL}{Natural Language}
	\acro{OSM}{OpenStreetMap}
	\acro{OWL}{Web Ontology Language}
	\acro{PFM}{Pseudo-F-Measures}
	\acro{PSO}{Particle-Swarm Optimization}
	\acro{QA}{Question Answering}
	\acro{RDF}{Resource Description Framework}
	\acro{SKOS}{Simple Knowledge Organization System}
	\acro{SPARQL}{SPARQL Protocol and RDF Query Language}
	\acro{SRL}{Statistical Relational Learning}
	\acro{SF}{surface forms}
	\acro{SW}{Semantic Web}
	\acro{UML}{Unified Modeling Language}
	\acro{WHO}{World Health Organization}
	\acro{WKT}{Well-Known Text}
	\acro{W3C}{World Wide Web Consortium}
	\acro{YPLL}{Years of Potential Life Lost}

	\acro{AOS}{Agricultural Ontology Services}
	\acro{AGRIS}{Agricultural Science and Technology}
	\acro{API}{Application Programming Interface}
	\acro{BPSO}{Binary Particle-Swarm Optimization}
	\acro{BPMLOD}{Best Practices for Multilingual Linked Open Data}
	\acro{CBD}{Concise Bounded Description}
	\acro{COG}{Content Oriented Guidelines}
	\acro{CSV}{Comma-Separated Values}
	\acro{CBMT}{Corpus-Based Machine Translation}
	\acro{CLIR}{Cross-Language Information Retrieval}
	\acro{DPSO}{Deterministic Particle-Swarm Optimization}
	\acro{DALY}{Disability Adjusted Life Year}

	\acro{ER}{Entity Resolution}
	\acro{EM}{Expectation Maximization}
	\acro{EBMT}{Example-Based Machine Translation}
	\acro{EBNF}{Extended Backus--Naur Form}
	\acro{EL}{Entity Linking}
	\acro{FAO}{Food and Agriculture Organization of the United Nations}
	\acro{GIS}{Geographic Information Systems}
	\acro{GHO}{Global Health Observatory}
	\acro{HDI}{Human Development Index}
	\acro{ICT}{Information and communication technologies}
    \acro{KB}{Knowledge Base}
	\acro{LR}  {Language Resource}
	\acro{LD}  {Linked Data}
	\acro{LLOD}  {Linguistic Linked Open Data}
	\acro{LIMES}{LInk discovery framework for MEtric Spaces}
	\acro{LS}  {Link Specifications}
	\acro{LDIF}{Linked Data Integration Framework}
	\acro{LGD} {LinkedGeoData}
	\acro{LOD} {Linked Open Data}
	\acro{MSE}{Mean Squared Error}
	\acro{MWE}{Multiword Expressions}
	\acro{MT}{Machine Translation}
	\acro{ML}{Machine Learning}
	\acro{NIF}{Natural Language Processing Interchange Format}
	\acro{NIF4OGGD}{NLP Interchange Format for Open German Governmental Data}
	\acro{NLP}{Natural Language Processing}
	\acro{NER}{Named Entity Recognition}
	\acro{NMT}{Neural Machine Translation}
	\acro{NN}{Neural Network}
	\acro{NLG}{Natural Language Generation}
	\acro{NED}{Named Entity Disambiguation}
	\acro{NERD}{Named Entity Recognition and Disambiguation}
	\acro{NL}{Natural Language}
	\acro{OSM}{OpenStreetMap}
	\acro{OWL}{Web Ontology Language}
	\acro{OOV}{out-of-vocabulary}
	\acro{PFM}{Pseudo-F-Measures}
	\acro{PSO}{Particle-Swarm Optimization}
	\acro{QA}{Question Answering}
	\acro{RDF}{Resource Description Framework}
	\acro{RBMT}{Rule-Based Machine Translation}
	\acro{SKOS}{Simple Knowledge Organization System}
	\acro{SPARQL}{SPARQL Protocol and RDF Query Language}
	\acro{SRL}{Statistical Relational Learning}
	\acro{SWT}{Semantic Web Technologies}
	\acro{SW}{Semantic Web}
	\acro{SMT}{Statistical Machine Translation}
	\acro{SWMT}{Semantic Web Machine Translation}

    \acro{TBMT} {Transfer-Based Machine Translation}
	\acro{UML}{Unified Modeling Language}
	\acro{WHO}{World Health Organization}
	\acro{WKT}{Well-Known Text}
	\acro{W3C}{World Wide Web Consortium}
	\acro{WSD}{Word Sense Disambiguation}
	\acro{YPLL}{Years of Potential Life Lost}

	\acro{AOS}{Agricultural Ontology Services}
	\acro{AGRIS}{Agricultural Science and Technology}
	\acro{API}{Application Programming Interface}
	\acro{A2KB}{Annotation to Knowledge Base}
	\acro{BPSO}{Binary Particle-Swarm Optimization}
	\acro{BPMLOD}{Best Practices for Multilingual Linked Open Data}
	\acro{BFS}{Breadth-First-Search}
	\acro{CBD}{Concise Bounded Description}
	\acro{COG}{Content Oriented Guidelines}
	\acro{CSV}{Comma-Separated Values}
	\acro{CCR}{cross-document co-reference}
	\acro{DPSO}{Deterministic Particle-Swarm Optimization}
	\acro{DALY}{Disability Adjusted Life Year}
	\acro{D2KB}{Disambiguation to Knowledge Base}

	\acro{ER}{Entity Resolution}
	\acro{EM}{Expectation Maximization}
	\acro{EL}{Entity Linking}
	\acro{FAO}{Food and Agriculture Organization of the United Nations}
	\acro{GIS}{Geographic Information Systems}
	\acro{GHO}{Global Health Observatory}
	\acro{HDI}{Human Development Index}
	\acro{ICT}{Information and communication technologies}
    \acro{KB}{Knowledge Base}
	\acro{LR}  {Language Resource}
	\acro{LD}  {Linked Data}
	\acro{LLOD}  {Linguistic Linked Open Data}
	\acro{LIMES}{LInk discovery framework for MEtric Spaces}
	\acro{LS}  {Link Specifications}
	\acro{LDIF}{Linked Data Integration Framework}
	\acro{LGD} {LinkedGeoData}
	\acro{LOD} {Linked Open Data}
	\acro{MSE}{Mean Squared Error}
	\acro{MWE}{Multiword Expressions}
	\acro{NIF}{NLP Interchange Format}
	\acro{NIF4OGGD}{NLP Interchange Format for Open German Governmental Data}
	\acro{NLP}{Natural Language Processing}
	\acro{NER}{Named Entity Recognition}
	\acro{NED}{Named Entity Disambiguation}
	\acro{NEL}{Named Entity Linking}
	\acro{NN}{Neural Network}
	\acro{OSM}{OpenStreetMap}
	\acro{OWL}{Web Ontology Language}
	\acro{PFM}{Pseudo-F-Measures}
	\acro{PSO}{Particle-Swarm Optimization}
	\acro{QA}{Question Answering}
	\acro{RDF}{Resource Description Framework}
	\acro{SKOS}{Simple Knowledge Organization System}
	\acro{SPARQL}{SPARQL Protocol and RDF Query Language}
	\acro{SRL}{Statistical Relational Learning}
	\acro{SF}{surface forms}
	\acro{UML}{Unified Modeling Language}
	\acro{WHO}{World Health Organization}
	\acro{WKT}{Well-Known Text}
	\acro{W3C}{World Wide Web Consortium}
	\acro{YPLL}{Years of Potential Life Lost}

 \acro{IRI}{Internationalized Resource Identifiers}
 \acro{URI}{Uniform Resource Identifier}
 \acro{RDF}{Resource Description Framework}
 \acro{OWL}{Web Ontology Language}
 \acro{MOS}{Manchester OWL Syntax}
 \acro{SPARQL}{SPARQL Query Language}
 
\end{acronym}  

\maketitle
\begin{abstract}
With the ever-growing generation of data for the Semantic Web comes an increasing demand for this data to be made available to non-semantic Web experts.
One way of achieving this goal is to translate the languages of the Semantic Web into natural language. 
We present LD2NL, a framework for verbalizing the three key languages of the Semantic Web, i.e., RDF, OWL, and SPARQL. 
Our framework is based on a bottom-up approach to verbalization.
We evaluated LD2NL in an open survey with 86 persons.
Our results suggest that our framework can generate verbalizations that are close to natural languages and that can be easily understood by non-experts.
Therewith, it enables non-domain experts to interpret Semantic Web data with more than 91\% of the accuracy of domain experts.
\end{abstract}

\section{Introduction}
\label{sec:intro}

\ac{NLG} is the process of automatically generating coherent \ac{NL} text from non-linguistic data~\cite{reiter2000building}. Recently, the field has seen an increased interest in the development of \ac{NLG} systems focusing on verbalizing resources from \ac{SW} data~\cite{gardent2017creating}.
The \ac{SW} aims to make information available on the Web easier to process for machines and humans. However, the languages underlying this vision, i.e., \ac{RDF}, \ac{SPARQL} and \ac{OWL}, are rather difficult to understand for non-expert users. For example, while the meaning of the OWL class expression \texttt{Class: Professor SubClassOf: worksAt SOME University} is obvious to every \ac{SW} expert, this expression (``Every professor works at a university'') is rather difficult to fathom for lay persons.

Previous works such as SPARQL2NL~\cite{sparql2nl} and SPARTIQULATION~\cite{spartiqulation} have already shown the usefulness of the verbalization of SPARQL~\footnote{SPARQL is the query language for RDF data.} and RDF in areas such as question answering~\cite{deqa} and the explanation of the output of systems based on\ac{SW} technologies~\cite{sparql2nldemo}. However, other \ac{SW} languages are rarely investigated, such as OWL.

In this paper, we present an open-source holistic \ac{NLG} framework for the \ac{SW}, named LD2NL, which facilitates the verbalization of the three key languages of the \ac{SW}, i.e., RDF, OWL, and SPARQL into \ac{NL}. Our framework is based on a bottom-up paradigm for verbalizing \ac{SW} data. Additionally, LD2NL builds upon \textit{SPARQL2NL} as it is open-source and the paradigm it follows can be reused and ported to RDF and OWL. Thus, LD2NL is capable of generating either a single sentence or a summary of a given resource, rule, or query. To validate our framework, we evaluated LD2NL using experts 66 in \ac{NLP} and \ac{SW} as well as 20 non-experts who were lay users or non-users of \ac{SW}. The results suggest that LD2NL generates texts which can be easily understood by humans. The version of LD2NL used in this paper, all experimental results will be publicly available.


\section{Related Work}
\label{sec:related_work}

According to~\newcite{gatt2017survey}, there has been a plenty of works which investigated the generation of \ac{NL} texts from \ac{SWT} as an input data~\cite{cimiano2013exploiting,duma2013generating,ell2014language,biran2015discourse}. However, the subject of research has only recently gained significant momentum due to the great number of published works in the WebNLG \cite{colin2016webnlg} challenge along with deep learning techniques~\cite{sleimi2016generating,mrabet2016aligning}. \ac{RDF} has also been showing promising benefits to the generation of benchmarks for evaluating \ac{NLG} systems~\cite{gardent2017creating,perez2016building}. 

Despite the plethora of recent works written on handling RDF data, only a few have exploited the generation of \ac{NL} from OWL and SPARQL. For instance, ~\newcite{Androutsopoulos2013} generates sentences in English and Greek from OWL ontologies. Also, SPARQL2NL~\cite{ngonga2013sorry} uses rules to verbalize atomic constructs and combine their verbalization into sentences.  Therefore, our goal with LD2NL is to provide a complete framework to verbalize \ac{SW} concepts rather than become the state of the art on the respective tasks.


\section{Background}
\label{sec:preliminaries}
\subsection{OWL}
\label{subsec:owl}
\ac{OWL}\footnote{\url{www.w3.org/TR/owl2-overview/}}~\cite{owl2-overview} is the de-facto standard for machine processable and interoperable ontologies on the \ac{SW}. In its second version, \ac{OWL} is equivalent to the description logic $\mathcal{SROIQ}(D)$. Such expressiveness has a higher computational cost but allows the development of interesting applications such as automated reasoning~\cite{buhmann2016dl}. OWL 2 ontologies consist of the following three different syntactic categories:

\textit{\textbf{Entities}}, such as \emph{classes}, \emph{properties}, and \emph{individuals}, are identified by IRIs.
 They form the primitive terms and constitute the basic elements of an ontology. 
 Classes denote sets of individuals and properties link two individuals or an individual and a data value along a property.
 For example, a class \class{:Animal} can be used to represent the set of all animals. 
 Similarly, the object property \property{:childOf} can be used to represent the parent-child relationship and the data property \property{:birthDate} assigns a particular birth date to 
an individual. 
 Finally, the individual \individual{:Alice} can be used to represent a particular person called "Alice".

\textit{\textbf{Expressions}} represent complex notions in the domain being described. For example, a \emph{class expression} describes a set of individuals in terms of the restrictions on the individuals' characteristics.
 OWL offers existential (\textbf{SOME}) or universal (\textbf{ONLY}) qualifiers and a variety of typical logical constructs, such as negation (\textbf{NOT}), other Boolean operators 
(\textbf{OR}, \textbf{AND}), and more constructs such as cardinality restriction (\textbf{MIN}, \textbf{MAX}, \textbf{EXACTLY}) and value restriction (\textbf{VALUE}), to create 
class expressions. 
Such constructs can be combined in arbitrarily complex class expressions \texttt{CE} according to the following grammar
\begin{lstlisting}[language=manchester,mathescape=true]
 CE = A | C AND D | C OR D | NOT C | R SOME C | R ONLY C | R MIN n | R MAX n | R EXACTLY n | R VALUE a | {a$_1$,...,a$_m$}
\end{lstlisting}
where \texttt{A} is an atomic class, \texttt{C} and \texttt{D} are class expressions, \texttt{R} is an object property,  \texttt{a} as well as \texttt{a}$_1$ to \texttt{a}$_m$ with 
$\texttt{m} \geq 1$ are individuals, and $\texttt{n} \geq 0$ is an integer.

\textit{\textbf{Axioms}} are statements that are asserted to be true in the domain being described. Usually, one distinguish between (1) \emph{terminological} and (2) \emph{assertional} axioms. (1) terminological axioms are used to describe the structure of the domain, i.e., the relationships between classes resp. class expressions. For example, using a subclass axiom (\textbf{SubClassOf:}), one can state that the class \class{:Koala} is a subclass of the class \class{:Animal}.
Classes can be subclasses of other classes, thus creating a taxonomy.
In addition, axioms can arrange properties in hierarchies (\textbf{SubPropertyOf:}) and can assign various characteristics (\textbf{Characteristics:}) such as transitivity or reflexivity to 
them. 
(2) Assertional axioms formulate facts about individuals, especially the classes they belong to and their mutual relationships. OWL can be expressed in various syntaxes with the most common computer readable syntax being RDF/XML
A more human-readable format is the \ac{MOS}~\cite{horridge2006}.
For example, the class expression that models people who work at a university that is located in Spain could be as follows in \ac{MOS}:
\begin{lstlisting}[language=manchester]
Person AND worksAt SOME (University AND locatedIn VALUE Spain)
\end{lstlisting}
Likewise, expressing that every professor works at a university would read as
\begin{lstlisting}[language=manchester]
Class: Professor
  SubClassOf: worksAt SOME University
\end{lstlisting} 

\subsection{RDF}
\label{subsec:rdf}
\ac{RDF}~\cite{rdf-concepts} uses a graph-based data model for representing knowledge. Statements in RDF are expressed as so-called triples of the form \triple{\normalsize subject, predicate, object}.
RDF subjects and predicates are \acp{IRI} and objects are either \acp{IRI} or literals.\footnote{For simplicity, we omit RDF blank nodes in subject or object position.}
RDF literals always have a datatype that defines its possible values.
A predicate denotes a \emph{property} and can also be seen as a binary relation taking subject and object as arguments.
For example, the following triple expresses that Albert Einstein was born in Ulm:
\begin{lstlisting}[language=ttl]
 :Albert_Einstein :birthPlace :Ulm .
\end{lstlisting}

\subsection{SPARQL}
\label{subsec:sparql}
Commonly, the selection of subsets of RDF is performed using the SPARQL query language.\footnote{\scriptsize\url{http://www.w3.org/TR/sparql11-query}}
\ac{SPARQL} can be used to express queries across diverse data sources. 
\emph{Query forms} contain variables that appear in a solution result. They can be used to select all or a subset of the variables bound in a pattern match. They exist in four different instantiations, i.e., \emph{SELECT}, \emph{CONSTRUCT}, \emph{ASK} and \emph{DESCRIBE}. The \emph{SELECT} query form is the most commonly used and is used to return rows of variable bindings. Therefore, we use this type of query in our explanation. 
\emph{CONSTRUCT} allows to create a new RDF graph or modify the existing one through substituting variables in a graph templates for each solution.
\emph{ASK} returns a Boolean value indicating whether the graph contains a match or not.
 Finally, \emph{DESCRIBE} is used to return all triples about the resources matching the query. 
For example, \ref{lst:query} represents the following query ``Return all scientists who were born in Ulm''.

\begin{lstlisting}[label=lst:query, float=htb, style=sparql, numbers=none, 
caption=All scientists who were born in Ulm]
SELECT ?person
WHERE { 
    ?person a dbo:Scientist;
        dbo:birthPlace dbr:Ulm. 
}
\end{lstlisting}
\vspace{-2mm}
\vspace{-2mm}

\section{LD2NL Framework}
\label{sec:approach}

The goal of LD2NL is to provide an integrated system which generates a complete and correct \ac{NL} representation for the most common used \ac{SW} modeling languages RDF and OWL, and SPARQL. 
In terms of the standard model of \ac{NL} generation proposed by Reiter \& Dale~\cite{REDA00}, our steps mainly play the role of the micro-planner, with focus on aggregation, lexicalization, referring expressions and linguistic realization. 
In the following, we present our approach to formalizing \ac{NL} sentences for each of the supported languages.

\subsection{From RDF to NL}
\label{sec:rdf_entities}

\subsubsection{Lexicalization}

The lexicalization of RDF triples must be able to deal with resources, classes, properties and literals.

\textit{\textbf{Classes and resources}} The lexicalization of classes and resources is carried out as follows: 
Given a URI $u$ we ask for the English label of $u$ using a SPARQL query.\footnote{Note that it could be any property which returns a \ac{NL} representation of the given URI, see 
\cite{ell2011}.} 
If such a label does not exist, we use either the fragment of $u$ (the string after \verb|#|) if it exists, else the string after the last occurrence of \verb|/|.
Finally this \ac{NL} representation is realized as a noun phrase, and in the case of classes is also pluralized.
As an example, \resource{:Person} is realized as \texttt{people} (its label).

\textit{\textbf{Properties}} The lexicalization of properties relies on the insight that most property labels are either nouns or verbs.
While the mapping of a particular property \texttt{p} can be unambiguous, 
some property labels are not as easy to categorize. 
For examples, the label \texttt{crosses} can either be the plural form of the noun \texttt{cross} or the third person singular present form of the verb \texttt{to cross}. 
To automatically determine which realization to use, we relied on the insight that the first and last word of a property label are often the key to determining the type of the 
property:
properties whose label begins with a verb (resp. noun or gerund) are most to be realized as verbs (resp. nouns).
We devised a set of rules to capture this behavior, which we omit due to space restrictions.
In some cases (such as \texttt{crosses}) none of the rules applied. 
In these cases, we compare the probability of $P(p|\texttt{noun})$ and $P(p|\texttt{verb})$ by measuring 
\begin{equation}
\footnotesize
P(p|X) = \frac{\sum\limits_{t \in \text{\em synset}(p|X)} \log_2(f(t))}{\sum\limits_{t' \in \text{\em synset}(p)} \log_2(f(t'))},
\end{equation}
where $\text{\em synset}(p)$ is the set of all synsets of $p$, $\text{\em synset}(p|X)$ is the set of all synsets of $p$ that are of the syntactic class $X  \in 
\{\texttt{noun},\texttt{verb}\}$ and $f(t)$ is the frequency of use of $p$ in the sense of the synset $t$ according to WordNet.
For 
\begin{equation}
\footnotesize
\frac{P(p|\texttt{verb})}{P(p|\texttt{noun})} \geq \theta,
\end{equation}
 we choose to realize \texttt{p} as a noun; else we realized it as a verb. 
For $\theta=1$, for example, \resource{dbo:crosses} is realized as a verb.

\textit{\textbf{Literals}} Literals in an RDF graph usually consist of a \emph{lexical form} \texttt{LF} and a \emph{datatype IRI} \texttt{DT}, represented as \literal{LF}{DT}.
Optionally, if the datatype is \rdfLangString, a non-empty \emph{language tag} is specified and the literal is denoted as \emph{language-tagged 
string}\footnote{In RDF 1.0 literals have been divided into 'plain' literals with no type and optional language tags, and typed literals.}. The realization of language-tagged strings is done by using simply the lexical form, while omitting the language tag.
For example, \texttt{"Albert Einstein"@en} is realized as \texttt{Albert Einstein}.
For other types of literals, we further differentiate between built-in and user-defined datatypes. 
For the former, we also use the lexical form, e.g. \literalPrefixed{123}{xsd:int} $\Rightarrow$ \texttt{123}, while the latter are processed by using the literal value with its representation of the datatype IRI, e.g., \literalPrefixed{123}{dt:squareKilometre} as \texttt{123 square kilometres}.

\subsubsection{Realizing single triples}
\label{sec:single_triple}
The realization $\rho$ of a triple \triple{\normalsize s p o} depends mostly on the verbalization of the predicate \texttt{p}.
If \texttt{p} can be realized as a noun phrase, then a possessive clause can be used to express the semantics of \triple{\normalsize s p o}, more formally
\begin{enumerate}
\footnotesize
  \item \texttt{$\rho$(s p o) $\Rightarrow$ poss($\rho$(p),$\rho$(s))$\,\wedge\,$subj(BE,$\rho$(p))$\,\\
  \wedge\,$dobj(BE,$\rho$(o))}
\end{enumerate}
For example, if $\rho(\texttt{p})$ is a relational noun like \texttt{birth place} e.g. in the triple \triple{:Albert\_Einstein :birthPlace :Ulm}, then the verbalization is \texttt{Albert 
Einstein's birth place is Ulm}.
Note that \texttt{BE} stands for the verb ``to be''. In case \texttt{p}'s realization is a verb, then the triple can be verbalized as follows:
\begin{enumerate}[resume]
\footnotesize
  \item \texttt{$\rho$(s p o) $\Rightarrow$ subj($\rho$(p),$\rho$(s))$\,\wedge\,$dobj($\rho$(p),$\rho$(o))}
\end{enumerate}
For example, in \triple{:Albert\_Einstein :influenced :Nathan\_Rosen} $\rho(\texttt{p})$ is the verb \texttt{influenced}, thus, the verbalization is \phrase{ \small Albert Einstein influenced Nathan Rosen}.

\subsection{Realization - RDF Triples to NL}
\label{sec:triples}

The same procedure of generating a single triple can be applied for the generation of each triple in a set of triples. However, the \ac{NL} output would contain redundant information and consequently sound very artificial. Thus, the goal is to transform the generated description to sound more natural. To this end, we focus on two types of transformation rules (cf.~\cite{dalianishovy}): \emph{ordering and clustering} and \emph{grouping}. In the following, we describe the transformation rules we employ in more detail. Note that clustering and ordering (\ref{subsec:ordering}) is applied before grouping (\ref{subsec:grouping}).

\subsubsection{Clustering and ordering rules} \label{subsec:ordering}
We process the input trees in descending order with respect to the frequency of the variables they contain, starting with the projection variables and only after that turning to other variables. As an example, consider the following triples about two of the most known people in the world:
\begin{lstlisting}[language=ttl]
:William_Shakespeare rdf:type :Writer .
:Albert_Einstein :birthPlace :Ulm .
:Albert_Einstein :deathPlace :Princeton
:Albert_Einstein rdf:type :Scientist .
:William_Shakespeare :deathDate 
"1616-04-23"^^xsd:date .
\end{lstlisting}

The five triples are verbalized as given in \ref{ex:triple1}--\ref{ex:triple5}. Clustering and ordering first take all sentences containing the subject \resource{:Albert\_Einstein}, i.e. \ref{ex:triple2} 
--\ref{ex:triple4}, which are ordered such that copulative sentences (such as {\sf \footnotesize Albert Einstein is a scientist}) come before 
other sentences, and then takes all sentences containing the remaining subject \resource{:William\_Shakespeare} in \ref{ex:triple1} and \ref{ex:triple5}
resulting in a sequence of sentences as in \ref{ex:triples}.
\begin{enumerate}[resume]
\item \begin{enumerate}
 \footnotesize
  \item \phrase{ \footnotesize William Shakespeare is a writer.} \label{ex:triple1}
  \item \phrase{ \footnotesize Albert Einstein's birth place is Ulm.} \label{ex:triple2}
  \item \phrase{ \footnotesize Albert Einstein's death place is Princeton.} \label{ex:triple3}
  \item \phrase{ \footnotesize Albert Einstein is a scientist.} \label{ex:triple4}
  \item \phrase{ \footnotesize William Shakespeare's death date is 23 April 1616.} \label{ex:triple5}
  \end{enumerate}
  \label{ex:sportsentences}
   \footnotesize
\item \phrase{ \footnotesize Albert Einstein is a scientist. Albert Einstein's birth place is Ulm. Albert Einstein's death place is Princeton. William Shakespeare's is a writer. William Shakespeare's 
death date is 23 April 1616.}
\label{ex:triples}
\end{enumerate}

\subsubsection{Grouping}\label{subsec:grouping}
\newcite{dalianishovy} describe grouping as a process ``collecting clauses with common elements and then collapsing the common elements''. The common elements are usually subject noun phrases and verb phrases (verbs 
together with object noun phrases), leading to \emph{subject grouping} and \emph{object grouping}. To maximize the grouping effects, we collapse common prefixes and suffixes of sentences, irrespective of whether they are full subject noun phrases or complete verb phrases. In the following we use $X_1,X_2,$\ldots$X_N$ as variables for the root nodes of the input sentences and $Y$ as variable for the root node of the output sentence. Furthermore, we abbreviate a subject $\texttt{subj}(X_i,s_i)$ as $\texttt{s}_i$, an object $\texttt{dobj}(X_i,o_i)$ as $\texttt{o}_i$, and a verb $\texttt{root}(\text{\em ROOT}_i,v_i)$ as $\texttt{v}_i$.

\textit{\textbf{Subject grouping}}
collapses the predicates (i.e. verb and object) of two sentences if their subjects are the same, as specified in \ref{rule:fcr} (abbreviations as above). 
\begin{enumerate}[resume]
\footnotesize
\item $\rho(\texttt{s}_1)=\rho(\texttt{s}_2)\wedge\texttt{cc}(\texttt{v}_1,\text{\em coord})$\\
      $\Rightarrow \texttt{root}(Y,\texttt{coord}(\texttt{v}_1,\texttt{v}_2))\wedge\, \texttt{subj}(\texttt{v}_1,\texttt{s}_1)\wedge
      \texttt{dobj}(\texttt{v}_1,\texttt{o}_1)\wedge \texttt{subj}(\texttt{v}_2,\texttt{s}_1)\wedge \texttt{dobj}(\texttt{v}_1,\texttt{o}_2) $
\label{rule:fcr}
\end{enumerate}
An example are the sentences given in \ref{ex:einstein}, which share the subject \phrase{ \footnotesize Albert Einstein} 
and thus can be collaped into a single sentence. 
\begin{enumerate}[resume]
 \footnotesize
\item \phrase{ \footnotesize Albert Einstein is a scientist and Albert Einstein is known for general relativity.} \\
      $\Rightarrow$
      \phrase{ \footnotesize Albert Einstein is a scientist and known for general relativity.} 
\label{ex:einstein}
\end{enumerate}
\textit{\textbf{Object grouping}}
collapses the subjects of two sentences if the realizations of the verbs and objects of the sentences are the same, 
where the $\text{\em coord}\in\{\text{\sf and},\text{\sf or}\}$ is the coordination combining the input sentences $X_1$ and $X_2$, 
and $\texttt{coord}\in\{\texttt{conj},\texttt{disj}\}$ is the corresponding coordination combining the subjects.
\begin{enumerate}[resume]
\footnotesize
\item $\rho(\texttt{o}_1)=\rho(\texttt{o}_2)\wedge\rho(\texttt{v}_1)=\rho(\texttt{v}_2)\wedge\texttt{cc}(\texttt{v}_1,\text{\em coord})$\\ 
$\Rightarrow \texttt{root}(Y,\texttt{PLURAL}(\texttt{v}_1))\wedge\, \texttt{subj}(\texttt{v}_1,\texttt{coord}(\texttt{s}_1,\texttt{s}_2))\wedge
\texttt{dobj}(\texttt{v}_1,\texttt{o}_1) $
\label{rule:bcr}
\end{enumerate}
For example, the sentences in \ref{ex:bostonborn} share their verb and object, thus they can be collapsed into a single sentence. Note that to this end the singular auxiliary {\sf \footnotesize was} needs to be transformed into its plural form {\sf \footnotesize were}.
\begin{enumerate}[resume]
\footnotesize
\item \phrase{ \footnotesize Benjamin Franklin was born in Boston. Leonard Nimoy was born in Boston.}
      $\Rightarrow$ \phrase{ \footnotesize Benjamin Franklin and Leonard Nimoy were born in Boston.} 
\label{ex:bostonborn}
\end{enumerate}




\subsection{From OWL to NL}\label{sec:owl}
OWL 2 ontologies consist of Entities, Expressions and Axioms as introduced in \autoref{subsec:owl}. While both expressions and axioms can be mapped to RDF\footnote{\url{http://bit.ly/2Mc0vIw}}, i.e. into a set of RDF triples, using this mapping and applying the 
triple-based verbalization on it would lead to a non-human understandable text in many cases. For example, the intersection of two classes \class{:A} and \class{:B} can be represented in RDF 
by the six triples
\begin{lstlisting}[language=ttl]
 _:x rdf:type owl:Class .
 _:x owl:intersectionOf _:y1 .
 _:y1 rdf:first :A .
 _:y1 rdf:rest _:y2 .
 _:y2 rdf:first :B .
 _:y2 rdf:rest rdf:nil .
\end{lstlisting}
The verbalization of these triples would result in \texttt{Something that is a class and the intersection of something whose first is A and whose rest is 
something whose first is B and whose rest ist nil.}, which is obviously far away from how a human would express it in \ac{NL}. Therefore, generating \ac{NL} from OWL requires a different procedure based on its syntactic categories, OWL expressions and OWL axioms. 
We show the general rules for each of them in the following.

\subsubsection{OWL Class Expressions}\label{subsec:owl_ce}
In theory, class expressions can be arbitrarily complex, but as it turned out in some previous analysis~\cite{power2010}, in practice they seldom arise and can be seen as some corner cases.
For example, an ontology could contain the following class expression about people and their birth place:
\begin{lstlisting}[language=manchester]
 Person AND birthPlace SOME (City AND locatedIn VALUE France)
\end{lstlisting}
Class expressions do have a tree-like structure and can simply be parsed into a tree by means of the binary OWL class expressions constructors contained in it. For our example, this 
would result in the following tree:

\begin{tikzpicture}
\tikzset{every tree node/.style={align=center,anchor=base,font=\scriptsize\ttfamily},
edge from parent/.style= 
            {thick, draw,
                edge from parent path={(\tikzparentnode.south)
                                        -- +(0,-8pt)
                                        -| (\tikzchildnode)}}
}
\tikzset{frontier/.style={distance from root=9\baselineskip}}
\Tree [.\textbf{AND} Person  [.\textbf{SOME} birthPlace [.\textbf{AND} City [.\textbf{VALUE} locatedIn France ] ] ] ]
\end{tikzpicture}

Such a tree can be traversed in post-order, i.e. sub-trees are processed before their parent nodes recursively.
For the sake of simplicity, we only process sub-trees that represent proper class expression in our example, i.e. we omit \texttt{birthPlace}, \texttt{locatedIn}, and \texttt{France}.
Moreover and again for simplicity, we'll explain the transformation process by starting from the right-hand side of the tree. Thus, in our example we begin with the class expression \texttt{City} which is transformed to \texttt{everything that is a city} and 
\texttt{locatedIn VALUE France}
resulting in \texttt{everything that is located in France} by application of a rule.
Both class expressions are used in the conjunction
\texttt{City AND locatedIn VALUE France}. 
Thus, the next step would be to merge both phrases.
An easy way is to use the coordinating conjunction \texttt{and}, i.e. \texttt{everything that is a city and everything that is located in France}.
Although the output of this transformation is correct, it still contains unnecessarily redundant information.
Therefore, we apply the aggregation procedure described in~\Cref{subsec:grouping}, i.e. we get \texttt{everything that is a city and located in France}.
Yet, the aggregation can still be improved: if there is any atomic class in the conjunction, we know that this is more specific than the placeholder \texttt{everything}.
Thus, we can replace it by the plural form of the class, finally resulting in \texttt{cities that are located in France}. The same procedure is applied for its parent class expression being the existential restriction
\begin{lstlisting}[language=manchester]
birthPlace SOME (City AND locatedIn VALUE France)
\end{lstlisting}

This will be transformed to \texttt{everything whose birth place is a city that is located in France}.
Note, that we used the singular form here, assuming that the property \texttt{birthPlace} is supposed to be functional in the ontology.
In the last step, we process the class expression
\texttt{Person},
which gives us \texttt{everything that is a person}.
Again, due to the conjunction we merge this result with with the previous one, such that in the end we get
\texttt{people whose birth place is a city that is located in France}.

\subsubsection{OWL Axioms}
As we described in~\Cref{sec:owl}, OWL axioms can roughly be categorized into terminological and assertional axioms.
Therefore, we have different procedures for processing each category:

\textit{\textbf{Assertional Axioms}} (ABox Axioms)\mbox{} - Most assertional axioms assert individuals to atomic classes or relate individuals to another individual resp. literal value.
For example, axioms about the type as well as birth place and birth date of Albert Einstein can be expressed by
\begin{lstlisting}[language=manchester]
Individual: Albert_Einstein
  Types: Person
  Facts: birthPlace Ulm, birthDate "1879-03-14"^^xsd:date 
\end{lstlisting} 
Those axioms can simply be rewritten as triples, thus, we can use the same procedure as we do for triples (\Cref{sec:single_triple}).  
Converting them into \ac{NL} gives us \texttt{Albert Einstein is a person whose birth place is Ulm and whose birth date is 14 March 1879.}
OWL also allows for assigning an individual to a complex class expression. In that case we'll use our conversion of OWL class expressions as described in~\Cref{subsec:owl_ce}.

\textit{\textbf{Terminological Axioms}} (TBox Axioms)\mbox{} - According to~\newcite{power2010}, most of the terminological axioms used in ontologies are subclass axioms.
By definition, subclass and superclass can be arbitrarily complex class expressions $\texttt{CE}_1$ and $\texttt{CE}_2$, i.e. 
\lstinline[language=manchester,mathescape=true]{CE$_1$ SubClassOf CE$_2$},
but in praxis it is quite often only used with atomic classes as subclass or even more simple with the superclass also beeing an atomic class.
Nevertheless, we support any kind of subclass axiom and all other logical OWL axioms in LD2NL.
For simplicity, we  outline here how we verbalize subclass axioms in LD2NL. The semantics of a subclass axiom denotes that every individual of the subclass also belongs to the superclass.
Thus, the verbalization seems to be relatively straightforward, i.e. we verbalize both class expressions and follow the template : \phrase{ \small every \realize{\texttt{CE}$_1$} is a \realize{\texttt{CE}$_2$}}. Obviously, this works pretty well for subclass axioms with atomic classes only.
For example, the axiom
\begin{lstlisting}[language=manchester]
Class: Scientist
  SubClassOf: Person
\end{lstlisting} 
is verbalized as \texttt{every scientist is a person.}


\subsection{From SPARQL to NL}
A SPARQL \texttt{SELECT} query can be regarded as consisting of three parts: (1) a \emph{body section} \texttt{B}, which describes all data that has to be retrieved, (2) an \emph{optional section} \texttt{O}, which describes the data items that can be retrieved by the query if they exist, and 
(3) a \emph{modifier section} \texttt{M}, which describes all solution sequences, modifiers and aggregates 
that are to be applied to the result of the previous two sections of the query.
Let \emph{Var} be the set of all variables that can be used in a SPARQL query. 
In addition, let $R$ be the set of all resources, $P$ the set of all properties and $L$ the set of all literals contained in the target knowledge base of the SPARQL queries at hand.
We call $x \in \text{\em Var}\, \cup R \cup P \cup L$ an \emph{atom}.
The basic components of the body of a SPARQL query are triple patterns $(\texttt{s}, \texttt{p}, \texttt{o}) \in (\text{\em Var}\, \cup R) \times (\text{\em Var}\, \cup P) \times (\text{\em 
Var}\, \cup R \cup L)$.
Let $W$ be the set of all words in the dictionary of our target language. 
We define the realization function $\rho: \text{\em Var}\, \cup R \cup P \cup L \rightarrow W^*$ as the function which maps each atom to a word or sequence of words from the dictionary. 
The extension of $\rho$ to all SPARQL constructs maps all atoms $x$ to their realization $\rho(x)$ and defines how these atomic realizations are to be combined.
We denote the extension of $\rho$ by the same label $\rho$ for the sake of simplicity.
We adopt a rule-based approach to achieve this goal, where the rules extending $\rho$ to all valid SPARQL constructs are expressed in a conjunctive manner. This means that for premises 
$P_1,\ldots,P_n$ and consequences $K_1,\ldots,K_m$ we write $P_1 \wedge \ldots \wedge P_n \Rightarrow K_1 \wedge \ldots \wedge K_m$. 
The premises and consequences are explicated by using an extension of the Stanford dependencies\footnote{For a complete description of the vocabulary, see \url{https://stanford.io/2EzMjmo}.}.

For example, a possessive dependency between two phrase elements $e_1$ and $e_2$ is represented as $\texttt{poss}(e_1,e_2)$.
For the sake of simplicity, we slightly deviate from the Stanford vocabulary by not treating the copula {\sf to be} as an auxiliary, 
but denoting it as \texttt{BE}. 
Moreover, we extend the vocabulary by the constructs \texttt{conj} and \texttt{disj} which denote the conjunction resp. disjunction of two phrase elements.
In addition, we sometimes reduce the construct $\texttt{subj(y,x)}\wedge \texttt{dobj(y,z)}$ to the triple \texttt{(x,y,z)} $\in W^3$.

\section{Experiments}
\label{sec:experiments}

We evaluated our approach in three different experiments based on human ratings. We divided the volunteers into two groups---domain experts and non-experts. The group of domain experts comprised 66 persons while there were 20 non-experts forming the second group. In the first experiment, an OWL axiom and its verbalization were shown to the experts who were asked to rate the verbalization regarding the two following measures according to \newcite{gardent2017creating}: (1) Adequacy: Does the text contain only and all the information from the data? (2) Fluency: Does the text sound fluent and natural?. For both measures the volunteers were asked to rate on a scale from 1 (Very Bad) to 5 (Very Good). The experiment was carried out using 41 axioms of the Koala ontology.\footnote{\url{https://bit.ly/2K8BWts}} Because of the complexity of OWL axioms, only domain experts were asked to perform this experiment.

In the second experiment, a set of triples describing a single resource and their verbalization were shown to the volunteers.
The experts were asked to rate the verbalization regarding adequacy, fluency and \textit{completeness}, i.e., whether all triples have been covered. The non-experts were only asked to rate the fluency. The experiment was carried out using 6 DBpedia resources.
In the third experiment, the verbalization of an OWL class and 5 resources were shown to the human raters. For non-experts, the resources have been verbalized as well, while for domain experts the resources were presented as triples. The task of the raters was to identify the resource that fits the class description and, thus, is an instance of the class. We used 4 different OWL axioms and measured the amount of correct identified class instances.

\begin{figure*}[htb]
 \centering
 \includegraphics[width=0.30\textwidth]{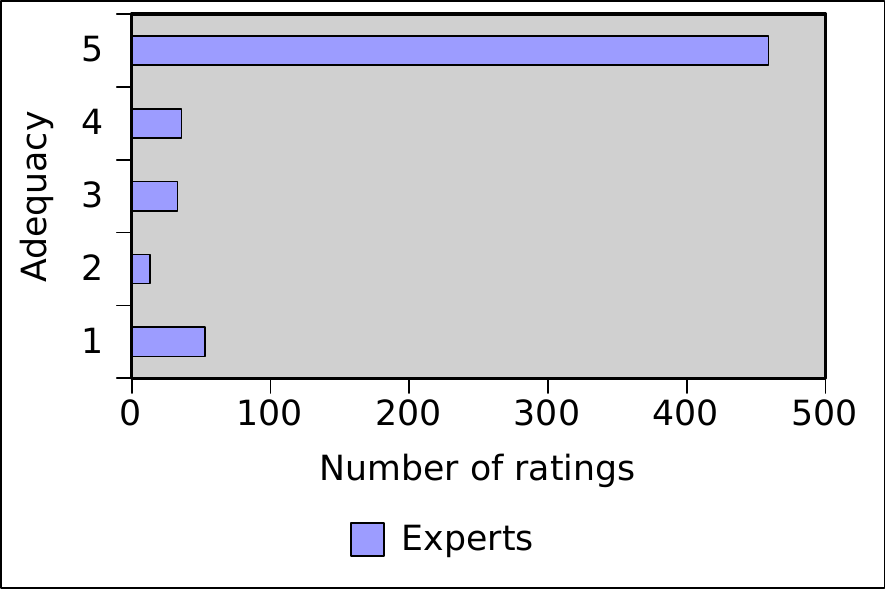} \quad
 \includegraphics[width=0.30\textwidth]{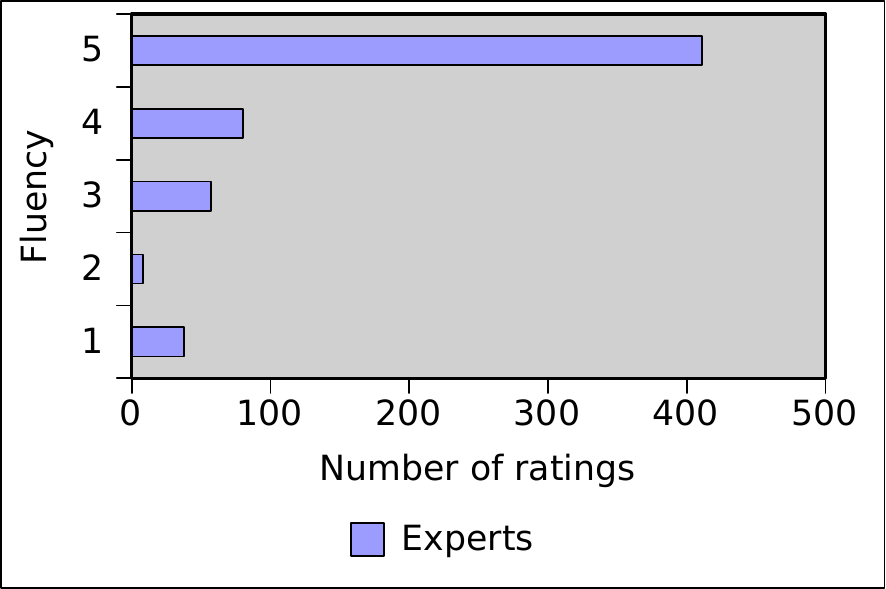}
 \caption{Experiment I: adequacy (left) and fluency (right) ratings}
 \label{fig:exp1}
\end{figure*}

\textit{\textbf{Results}} In our first series of experiments, the verbalization of OWL axioms, we achieved an average adequacy of 4.4 while the fluency reached 4.38. In addition, more than 77\% of the verbalizations were assigned the maximal adequacy (i.e., were assigned a score of 5, see Fig. \ref{fig:exp1}). The maximal score for fluency was achieved in more than 69\% of the cases (see Fig. \ref{fig:exp1}). This clearly indicates that the verbalization of axioms generated by LD2NL can be easily understood by domain experts and contains all the information necessary to access the input OWL class expression. 

\begin{figure*}[htb]
 \centering
 \includegraphics[width=0.30\textwidth]{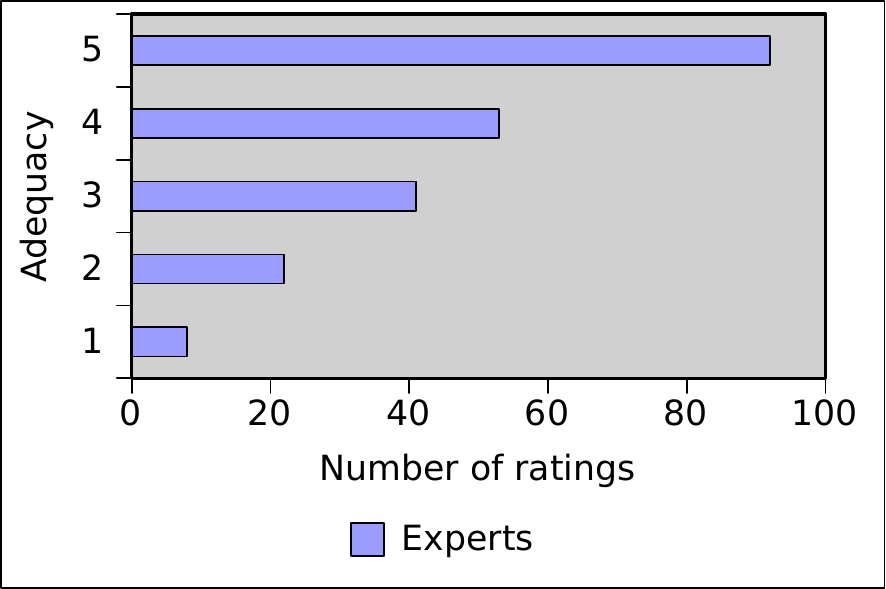} \quad
 \includegraphics[width=0.30\textwidth]{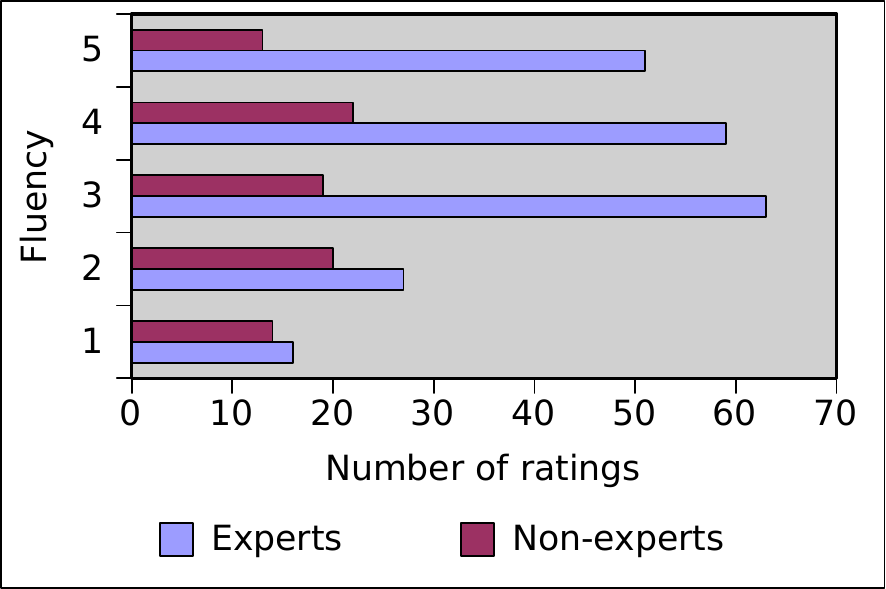}
 \quad
 \includegraphics[width=0.30\textwidth]{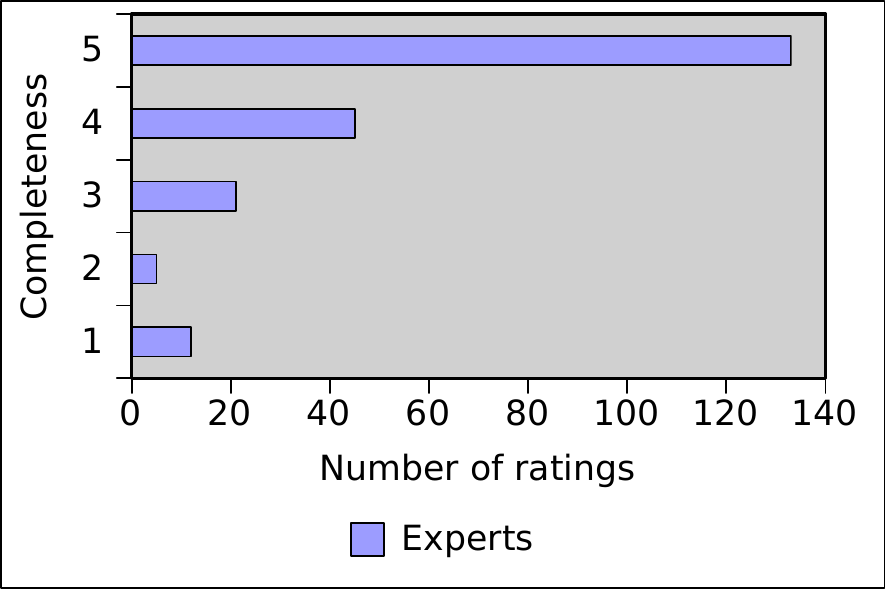}
 \caption{Experiment II: adequacy (left), fluency (middle) and completeness (left) results}
 \label{fig:exp2}
\end{figure*}

Experiments on the verbalization of summaries for RDF resources revealed that verbalizing resource summaries is a more difficult task. While the adequacy of the verbalization was assigned an average score of  3.92 by experts (see Fig. \ref{fig:exp2}), the fluency was assigned a average score of 3.47 by experts and 3.0 by non-experts (see Fig. \ref{fig:exp2}). What these results suggest is that (1) our framework generates sentences that are close to that which a domain expert would also generate (adequacy). However (2) while the sentence is grammatically sufficient for the experts, it is regarded by non-domain experts (which were mostly linguists, i.e., the worst-case scenario for such an evaluation) as being grammatically passably good but still worthy of improvement. The completeness rating achieves a score of 4.31 on average (see Fig. \ref{fig:exp2}). This was to be expected as we introduced a rule to shorten the description of resources that contain more than 5 triples which share a common subject and predicate. Finally, we measured how well the users and experts were able to understand the meaning of the text generated by our approach. As expected, the domain experts outperform the non-expert users by being able to find the answers to 87.2\% of the questions. The score achieved by non-domain experts, i.e., 80\%, still suggest that our framework is able to bridge the gap pertaining to understand RDF and OWL for non-experts from 0\% to 80\%, which is more than 91.8\% of the performance of experts.



\textit{\textbf{Discussion}}
Our evaluation results suggest that the verbalization of these languages is a non-trivial task that can be approached by using a bottom-up approach. As expected, the verbalization of short expressions leads to sentences which read as if they have been generated by a human. However, due to the complexity of the semantics that can be expressed by the languages at hand, long expressions can sound mildly artificial. Our results however also suggest that although the text generated can sound artificial, it is still clear enough to enable non-expert users to achieve results that are comparable to those achieved by experts. Hence, our first conclusion is that our framework clearly serves its purpose. Still, potential improvements can be derived from the results achieved during the experiments. In particular, we will consider the used of attention-based encoder-decoder networks to improve the fluency of complex sentences. 

\section{Conclusion and Future Work}
\label{sec:conclusion}

In this paper, we presented LD2NL, a framework for verbalizing \ac{SW} languages, especially on RDF and OWL while including the SPARQL verbalization provided by SPARQL2NL. Our evaluation with 86 persons revealed that our framework generates \ac{NL} that can be understood by lay users. While the OWL verbalization was close to \ac{NL}, the RDF was less natural but still sufficient to convey the meaning expressed by the corresponding set of triples. In future work, we aim to extend LD2NL to verbalize the languages SWRL~\cite{swrl} and SHACL~\cite{knublauch2017shapes}.

\section*{Acknowledgments} 	
This work was supported by the German Federal Ministry of Transport and Digital Infrastructure (BMVI) through the projects LIMBO (no. 19F2029I) and OPAL (no. 19F2028A). This work was supported by the German Federal Ministry of Economics and Technology (BMWI) in the projects RAKI (no. 01MD19012D) as well as by the BMBF project SOLIDE (no. 13N14456).

\bibliography{acl2017}
\bibliographystyle{acl_natbib}

\end{document}